\documentclass[nohyperref]{article}

\usepackage{microtype}
\usepackage{graphicx}
\usepackage{subfigure}
\usepackage{booktabs} 

\usepackage{hyperref}

\usepackage[accepted]{icml2022}

\usepackage{amsmath}
\usepackage{amssymb}
\usepackage{mathtools}
\usepackage{amsthm}
\usepackage{multirow}

\def\Vec#1{{\boldsymbol{#1}}}

\usepackage[capitalize,noabbrev]{cleveref}

\theoremstyle{plain}

\theoremstyle{definition}

\theoremstyle{remark}

\usepackage[textsize=tiny]{todonotes}

\icmltitlerunning{Multi-Grained Vision Language Pre-Training: Aligning Texts with Visual Concepts}

\begin{document}

\twocolumn[
\icmltitle{Multi-Grained Vision Language Pre-Training: \\ Aligning Texts with Visual Concepts}



\icmlsetsymbol{equal}{*}

                \begin{icmlauthorlist}
                \icmlauthor{Yan Zeng}{bytedance}
                \icmlauthor{Xinsong Zhang}{bytedance}
                \icmlauthor{Hang Li}{bytedance}
                \end{icmlauthorlist}
                
                \icmlaffiliation{bytedance}{ByteDance AI Lab}
                \icmlcorrespondingauthor{Yan Zeng}{zengyan.yanne@bytedance.com}

\icmlkeywords{X-VLM, Vision Language Model, ICML}

\vskip 0.3in
]



\printAffiliationsAndNotice{}  

\begin{abstract}
Most existing methods in vision language pre-training rely on object-centric features extracted through object detection and make fine-grained alignments between the extracted features and texts. It is challenging for these methods to learn relations among multiple objects. To this end, we propose a new method called X-VLM~\footnote{The code and pre-trained models are available at \url{https://github.com/zengyan-97/X-VLM}.} to perform `multi-grained vision language pre-training.' The key to learning multi-grained alignments is to locate visual concepts in the image given the associated texts, and in the meantime align the texts with the visual concepts, where the alignments are in multi-granularity. Experimental results show that X-VLM effectively leverages the learned multi-grained alignments to many downstream vision language tasks and consistently outperforms state-of-the-art methods.
\end{abstract}

\section{Introduction}
Vision language pre-training aims to learn vision language alignments from a large number of image-text pairs. A pre-trained Vision Language Model (VLM) fine-tuned with a small amount of labeled data has shown the state-of-the-art performances in many Vision Language (V+L) tasks such as visual question answering and image-text retrieval.

\begin{figure}[ht]
\begin{center}
\centerline{\includegraphics[width=\columnwidth]{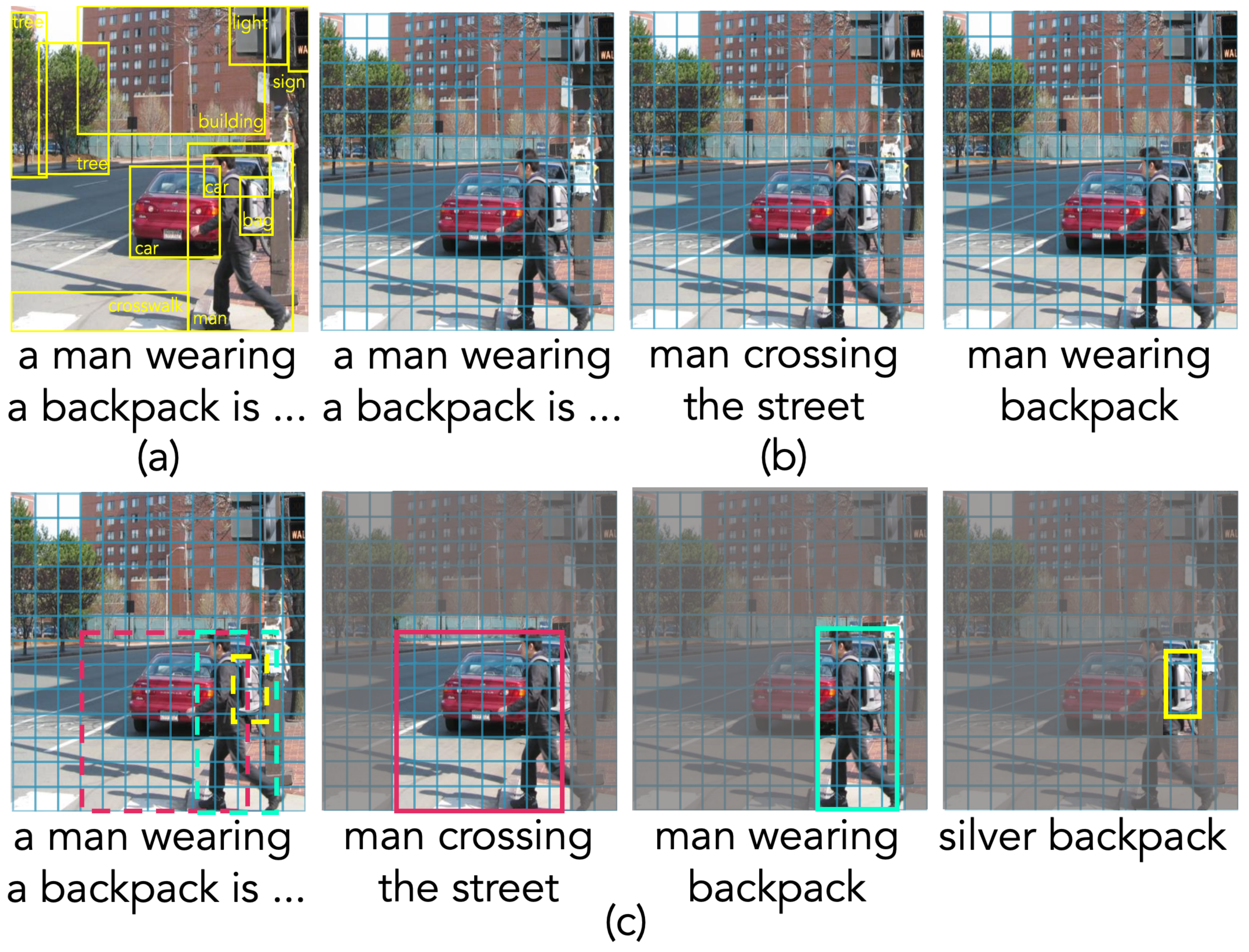}}
\caption{
A comparison of (a) the existing methods relying on object detection, (b) the methods aligning the texts with the whole image, and (c) our approach. 
}
\label{Fig:intro}
\end{center}
\end{figure}

Existing methods learning vision language alignments fall into two approaches as shown in Figure \ref{Fig:intro} (a, b). Most of them detect objects in the image and align the text with fine-grained (object-centric) features. They either utilize pre-trained object detectors~\cite{tan2019lxmert, lu2019vilbert, li2019visualbert, li2020unicoder, chen2020uniter, li2020oscar, gan2020large} or conduct object detection on-the-fly in the pre-training process~\cite{su2019vl, xu2021e2e}. The other methods do not rely on object detection and only learn alignments between the texts and coarse-grained (overall) features of the image~\cite{huang2020pixel, huang2021seeing, kim2021vilt, li2021align}.

Both the fine-grained and coarse-grained approaches have drawbacks. Object detection identifies all possible objects in the image, and some of them might not be relevant to the text. Object-centric features cannot easily represent relations among multiple objects, e.g. `` man crossing the street''. Moreover, it is challenging to pre-define the categories of objects suitable for downstream tasks. On the other hand, the coarse-grained approaches cannot effectively learn fine-grained alignments between vision and language, e.g. object-level, which has shown to be critical for some downstream tasks such as visual reasoning, visual grounding, and image captioning.

Ideally, we want a VLM to learn multi-grained alignments between vision and language in pre-training, which are not restricted to object-level or image-level, and leverage the learned alignments to downstream V+L tasks. Unfortunately, existing methods cannot satisfactorily handle multi-grained alignments between vision and language.

In this paper, we propose performing multi-grained vision language pre-training by aligning text descriptions with the corresponding visual concepts in images. Taking Figure \ref{Fig:intro} as an example, we have the following data for training: 1) the image caption describing the whole image; 2) region annotations such as ``man wearing backpack'' each of which has been related to a region in the image, while previous approaches roughly align the region descriptions with the whole image; 3) object labels such as ``backpack'' which are utilized by previous methods to train object detectors. We re-formulate the data, so that an image may have multiple bounding boxes, and a text \footnote{We take the object labels as text descriptions of objects.} is directly associated with the visual concept in each box. The `visual concept'~\cite{krishna2016visual, zhang2021vinvl, changpinyo2021conceptual} may be an object, a region, or the image itself, as the example in Figure~\ref{Fig:intro} (c). By doing so, our approach learns unlimited visual concepts associated with diverse text descriptions, which are also not restricted to object-level or image-level.

Our multi-grained model, denoted as X-VLM, consists of an image encoder that produces representations of visual concepts (including the image itself) in an image, a text encoder, and a cross-modal encoder that conducts cross-attention between the vision features and language features to learn vision language alignments. The key to learning multi-grained alignments is to optimize X-VLM by: 1) locating visual concepts in the image given associated texts by a combination of box regression loss and intersection over union loss; 2) in the meantime aligning the texts with the visual concepts, e.g. by a contrastive loss, a matching loss, and a masked language modeling loss, where the alignments are in multi-granularity, as illustrated in Figure~\ref{Fig:intro} (c). In fine-tuning and inference, X-VLM can leverage the learned multi-grained alignments to perform the downstream V+L tasks without bounding box annotations in the input images.

We demonstrate the effectiveness of our approach on various downstream tasks. On image-text retrieval, X-VLM learning multi-grained vision language alignments outperforms VinVL~\cite{zhang2021vinvl} which is based on object-centric features, achieving an absolute gain of 4.65\% in terms of R@1 score on MSCOCO. X-VLM also outperforms ALIGN~\cite{jia2021scaling}, ALBEF~\cite{li2021align}, and METER~\cite{dou2021empirical} by a large margin even though they are pre-trained on more data or have more parameters. On visual reasoning tasks, X-VLM achieves absolute improvements of 0.79\% on VQA and 1.06\% on NLVR2 compared to VinVL~\cite{zhang2021vinvl}, with a much faster inference speed. X-VLM also outperforms SimVLM$_\mathrm{base}$~\cite{wang2021simvlm} pre-trained with 1.8B in-house data, especially on NLVR2 by 2.4\%.  On visual grounding (RefCOCO+), X-VLM achieves absolute improvements of $4.5\%$ compared to UNITER~\cite{chen2020uniter} and $1.1\%$ compared to MDETR~\cite{kamath2021mdetr} which is specialized for grounding tasks. X-VLM also has comparable performance with SimVLM$_\mathrm{base}$ in the image caption generation task.

The contributions of this work are as follows:
\begin{itemize}

\item We propose performing multi-grained vision language pre-training to handle the alignments between texts and visual concepts. 

\item We propose to optimize the model (X-VLM) by locating visual concepts in the image given the associated texts and in the meantime aligning the texts with the visual concepts, where the alignments are in multi-granularity.

\item  We empirically verify that our approach effectively leverages the learned multi-grained alignments in fine-tuning. X-VLM consistently outperforms existing state-of-the-art methods on many downstream V+L tasks.  
\end{itemize}

\begin{figure*}[ht]
\begin{center}
\centerline{\includegraphics[width=2\columnwidth]{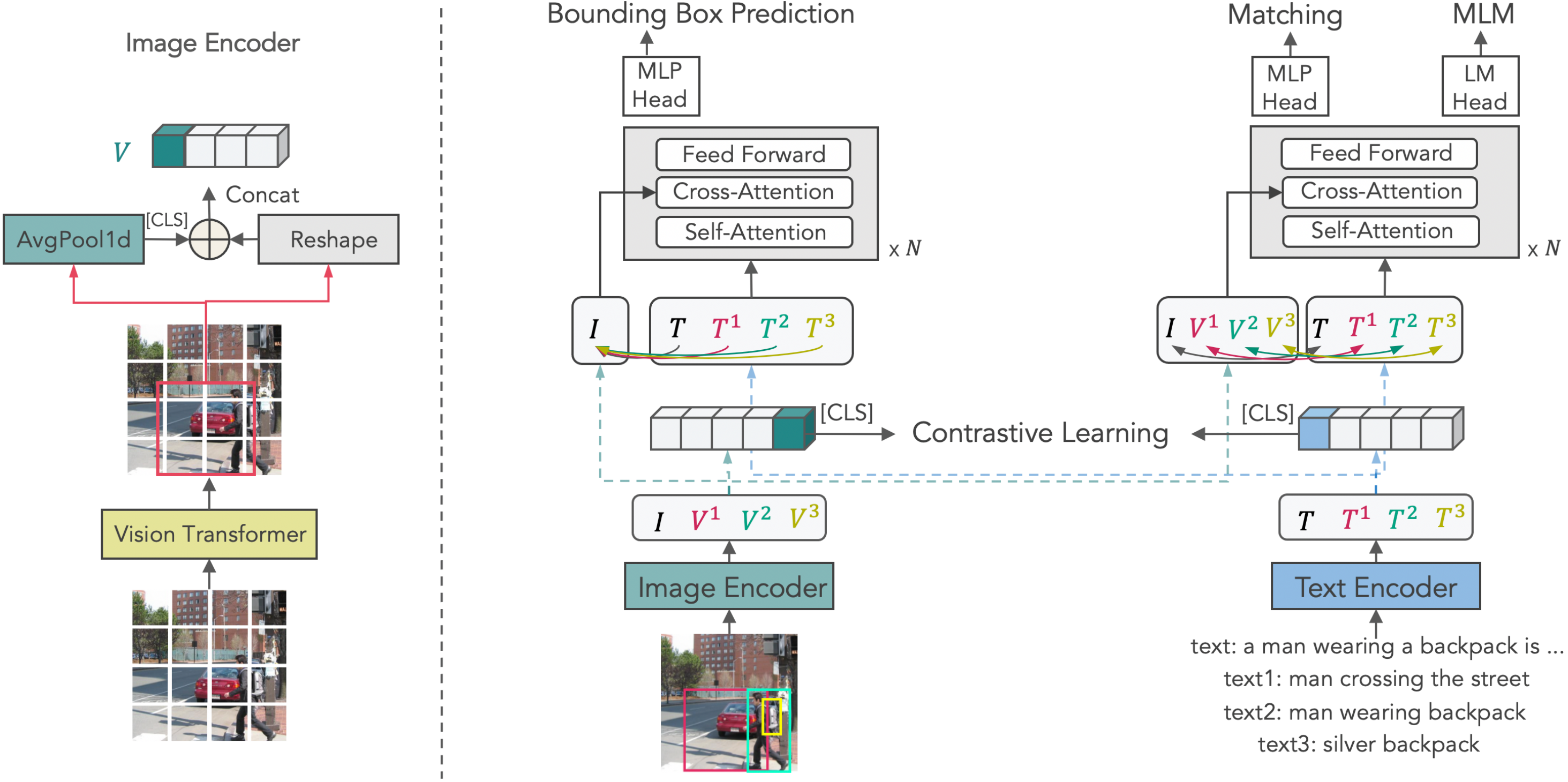}}
\caption{Pre-training model architecture and objectives of X-VLM. As shown on the left side, we extract features from the subset of patches from the vision transformer to represent images/regions/objects ($I$ and $V^{1-3}$), which are then paired with corresponding text features ($T$ and $T^{1-3}$) for contrastive learning, matching, and MLM. Meanwhile, the image ($I$) is paired with different textual descriptions ($T$ and $T^{1-3}$) for bounding box prediction to locate visual concepts in the image.}
\label{Fig:model}
\end{center}
\end{figure*}

\section{Related Work}
\label{sec:related}
The existing work on vision language pre-training typically falls into two categories: fine-grained and coarse-grained. 

Most existing methods belong to the fine-grained approach, which relies on object detection~\cite{tan2019lxmert, lu2019vilbert, li2019visualbert, li2020unicoder, chen2020uniter, li2020oscar, gan2020large, li2020unimo}. An object detector first identifies all regions that probably contain an object, then conducts object classification on each region. An image is then represented by dozens of object-centric features of the identified regions. Object detectors, such as Faster R-CNN~\cite{ren2015faster}, Bottom-Up and Top-Down Attention (BUTD)~\cite{anderson2018bottom}, are trained on image annotations of common objects, e.g. COCO~\cite{lin2014microsoft} (110K images) and Visual Genome~\cite{krishna2016visual} (100K), and can be utilized. VinVL~\cite{zhang2021vinvl} has, for example, achieved SoTA performances on many V+L tasks by utilizing a powerful object detector pre-trained with a large collection of image annotations (2.5M images). The challenge with the approach is that object-centric features cannot represent relations among multiple objects in multiple regions. Furthermore, it is not easy to define the categories of objects in advance that are useful for downstream V+L tasks.

The coarse-grained approach builds VLMs by extracting and encoding overall image features with convolutional network~\cite{jiang2020defense, huang2020pixel, huang2021seeing} or vision transformer~\cite{kim2021vilt, li2021align}. The performances are usually not as good as the fine-grained approach. Though object-centric features are only related to certain objects, learning fine-grained alignments, e.g. object-level, has shown to be critical for some downstream tasks such as visual reasoning and visual grounding. To cope with the problem, SOHO~\cite{huang2021seeing} employs online clustering on image features to obtain more comprehensive representations, ViLT~\cite{kim2021vilt} uses a more advanced vision transformer, i.e. Swin-Transformer~\cite{liu2021swin} for image encoding, and ALBEF~\cite{li2021align} exploits contrastive learning and momentum distillation in learning of image-text alignments. However, the improvements still cannot close the gap with the fine-grained approach.

Recently, there emerge some methods managing to learn both object-level and image-level alignments. However, these approaches still rely on object detectors and thus suffer from the aforementioned problems. For example, VL-BERT~\cite{su2019vl} incorporates Faster R-CNN into pre-training. E2E-VLP~\cite{xu2021e2e} adds an end-to-end object detection module (i.e. DETR~\cite{carion2020end}). Uni-EDEN~\cite{li2022uni} uses Faster R-CNN as the vision backbone. KD-VLP~\cite{liu2021kd} relies on external object detectors to perform object knowledge distillation. In contrast, X-VLM does not rely on object detection. Besides, X-VLM learns multi-grained vision language alignments, which are not restricted to object-level or image-level. Also, unlike Uni-EDEN, which aligns objects to language by object classification and aligns images to language by caption generation, X-VLM learns visual concepts in different granularities in a unified way. We will show the effectiveness of X-VLM in the experiments.

\section{Method}

\subsection{Overview}
X-VLM consists of an image encoder ($I_\mathrm{trans}$), a text encoder ($T_\mathrm{trans}$), and a cross-modal encoder ($X_\mathrm{trans}$). All encoders are based on Transformer~\cite{vaswani2017attention}. The cross-modal encoder fuses the vision features with the language features through cross-attention at each layer.

We re-formulate the widely used pre-training datasets (see Section \ref{sec:pretraindata}) so that an image may have multiple bounding boxes, and each of them is associated with a text that describes an object or a region, denoted as $(I, T, \{(V^{j}, T^j)\}^{N})$. Note that some images do not have associated texts, i.e., $T$ is NaN, and some images do not have bounding boxes, i.e., $N=0$. Here, $V^{j}$ is an object or region in the bounding box $\Vec{b}^{j}=(cx, cy, w, h)$ represented by the normalized center coordinates, width, and height of the box. When the image itself represents a visual concept, $\Vec{b} = (0.5, 0.5, 1, 1)$. Figure \ref{Fig:model} illustrates the architecture and pre-training objectives of X-VLM.

\subsection{Vision Encoding}  
The image encoder efficiently produces multi-grained visual concept representations in an image. The encoder is based on vision transformer~\cite{dosovitskiy2020image}. It first splits an image into non-overlapping patches and linearly embeds all patches. Then, these patches are passed into the transformer layers, yielding $\{\Vec{v}_1,...,\Vec{v}_{N^I}\}$. For an image of resolution of 224x224 and patch size of 32x32, we have $N^I=49$.

We assume that $\Vec{v}_{p_i}$ encodes the information of the corresponding patch $p_i$. Therefore, we represent a visual concept $V^{j}$ (object, region, or the image) that corresponds to a set of patches by aggregating information among the patches as shown in Figure~\ref{Fig:model}. Specifically, we reshape the patch features while keeping their position information, denoted as $\{\Vec{v}_{p^j_1},...,\Vec{v}_{p^j_{M}}\}$. $\{p^j_1,..., p^j_{M}\}$ are patches of $V^{j}$. We also calculate the average of the features to represent the whole visual concept, denoted as $\Vec{v}^j_\mathrm{cls}$, and prepend it.

The image encoder then creates $N+1$ concept representations in different granularities, represented as $I_\mathrm{trans}(V^j)=\{\Vec{v}^j_\mathrm{cls},\Vec{v}_{p^j_1},...,\Vec{v}_{p^j_{M}}\}, j\in[0, N]$. We let $I_\mathrm{trans}(V^0)$ denote the image representation in which all patch features are utilized. In the following section, we will describe how the representations are utilized in the learning of multi-grained alignments.

\subsection{Cross-Modal Modeling}

As shown in Figure~\ref{Fig:model}, we optimize X-VLM by locating visual concepts in the image given the corresponding texts and in the meantime aligning the texts and visual concepts, where the alignments are in multi-granularity.

\noindent\textbf{Bounding Box Prediction} We let the model predict the bounding box $\Vec{b}^{j}$ of visual concept $V^j$ given the image representation and the text representation, where $\Vec{b}^{j}=(cx, cy, w, h)$. By locating different visual concepts in the same image, we expect that the model better learns fine-grained vision language alignments. The bounding box is predicted by: 
\begin{equation}
\hat{\Vec{b}}^{j}(I, T^{j}) = \mathrm{Sigmoid}(\mathrm{MLP}(\Vec{x}^{j}_\mathrm{cls})),
\end{equation}
where Sigmoid is for normalization, MLP denotes multi-layer perceptron, and $\Vec{x}^{j}_\mathrm{cls}$ is the output \texttt{[CLS]} embedding of the cross-modal encoder given $I$ and $T^{j}$.

For bounding box prediction, $\ell_1$ is the most commonly-used loss. However, it has different scales for small and large boxes, even if their relative errors are similar. To mitigate this issue, we use a linear combination of the $\ell_1$ loss and the generalized Intersection over Union (IoU) loss~\cite{rezatofighi2019generalized}, which is scale-invariant. The overall loss is defined as: 
\begin{equation}
\mathcal{L}_\mathrm{bbox} = \mathbb{E}_{(V^{j}, T^{j}) \sim I; I \sim D} [\mathcal{L}_\mathrm{iou}(\Vec{b}_{j}, \hat{\Vec{b}}_{j}) + ||\Vec{b}_{j}- \hat{\Vec{b}}_{j}||_1 ]
\end{equation}

Meanwhile, we align texts and visual concepts by three objectives which are widely used in vision language pre-training~\cite{chen2020uniter, radford2021learning, li2021align}. We extend the objectives to incorporate multi-grained visual concepts in the images.

\noindent\textbf{Contrastive Learning} We predict (visual concept, text) pairs, denoted $(V,T)$, from in-batch negatives. Note that visual concepts include objects, regions, and images. Similar to \citet{radford2021learning}, we randomly sample a mini-batch of $N$ pairs, and calculate the in-batch vision-to-text similarity and text-to-vision similarity.

Given a pair $(V,T)$, $T$ is the positive example for $V$, and we treat the other $(N-1)$ texts within the mini-batch as negative examples. We define cosine similarity $s(V,T) = g_v(\Vec{v}_\mathrm{cls})^\top g_w(\Vec{w}_\mathrm{cls})$. $\Vec{w}_\mathrm{cls}$ is the output \texttt{[CLS]} embedding of the text encoder. $g_v$ and $g_w$ are transformations that map the \texttt{[CLS]} embeddings to normalized lower-dimensional representations. Then, we calculate the in-batch vision-to-text similarity as: 
\begin{equation}
p^\mathrm{v2t}(V) = \frac{\exp (s(V,T) / \tau)}{\sum_{i=1}^N \exp (s(V,T^i)/ \tau)},
\label{eq:pi2t}
\end{equation}

Similarly, the text-to-vision similarity is: 
\begin{equation}
p^\mathrm{t2v}(T) = \frac{\exp (s(V,T)/ \tau)}{\sum_{i=1}^N \exp (s(V^i,T)/ \tau)},
\end{equation}
where $\tau$ is a learnable temperature parameter. Let $\Vec{y}^\mathrm{v2t}(V)$ and $\Vec{y}^\mathrm{t2v}(T)$ denote the ground-truth one-hot similarity, in which only the positive pair has the probability of one. The contrastive loss is defined as the cross-entropy $\mathrm{H}$ between $\Vec{p}$ and $\Vec{y}$: 
\begin{align}
\label{eqn:itc}
\mathcal{L}_\mathrm{cl} = \frac{1}{2} \mathbb{E}_{V,T\sim D} \big[ & \mathrm{H}(\Vec{y}^\mathrm{v2t}(V),\Vec{p}^\mathrm{v2t}(V)) \notag \\
& + \mathrm{H}(\Vec{y}^\mathrm{t2v}(T),\Vec{p}^\mathrm{t2v}(T)) \big]
\end{align}

\noindent\textbf{Matching Prediction} We determine whether a pair of visual concept and text is matched. For each visual concept in a mini-batch, we sample an in-batch hard negative text by following $p^\mathrm{v2t}(V)$ in Equation~\ref{eq:pi2t}. Texts that are more relevant to the concept are more likely to be sampled. We also sample one hard negative visual concept for each text. We use $\Vec{x}_\mathrm{cls}$, the output \texttt{[CLS]} embedding of the cross-modal encoder, to predict the matching probability $p^\mathrm{match}$, and the loss is:
\begin{equation}
\label{eqn:itm}
\mathcal{L}_\mathrm{match} = \mathbb{E}_{V,T\sim D} \mathrm{H} (\Vec{y}^\textrm{match}, \Vec{p}^\textrm{match}(V,T))
\end{equation}
where $\Vec{y}^\textrm{match}$ is a 2-dimensional one-hot vector representing the ground-truth label.

\noindent\textbf{Masked Language Modeling} We predict the masked words in the text based on the visual concept. We randomly mask out the input tokens with a probability of 25\%, and the replacements are 10\% random tokens, 10\% unchanged, and 80\% \texttt{[MASK]}. We use the cross-modal encoder's outputs, and append a linear layer followed by softmax for prediction. Let $\hat{T}$ denote a masked text, and $\Vec{p}^j(V,\hat{T})$ denote the predicted probability of the masked token $t_j$. We minimize the cross-entropy loss:
\begin{equation}
\label{eqn:mlm}
\mathcal{L}_\mathrm{mlm} = \mathbb{E}_{t_j \sim \hat{T}; (V,\hat{T})\sim D} \mathrm{H} (\Vec{y}^j, \Vec{p}^j(V,\hat{T}))
\end{equation}
where $\Vec{y}^j$ is a one-hot distribution in which the ground-truth token $t_j$ has the probability of one.

Finally, the pre-training objective of X-VLM is defined as:
\begin{equation}
\mathcal{L} = \mathcal{L}_\mathrm{bbox} + \mathcal{L}_\mathrm{cl} + \mathcal{L}_\mathrm{match} + \mathcal{L}_\mathrm{mlm} 
\end{equation}

\section{Experiment}

\subsection{Pre-training Datasets}
\label{sec:pretraindata}

\begin{table}[ht]
	\caption
	{
		Statistics of the pre-training datasets. See Appendix \ref{app:ann} for detailed statistics of object and region annotations. 
	}
	\label{tbl:data}
    \small
	\centering	
	\begin{tabular}	{l | l | l |  l | l }
	\toprule
	 & Dataset & \# Images  & \# Captions & \# Ann \\
\midrule
	 \multirow{4}{*}{4M} & COCO & 0.11M & 0.55M & 0.45M \\
	 	 & VG & 0.10M & - & 5.7M \\
	 & SBU & 0.86M & 0.86M & -\\
	 & CC-3M & 2.9M & 2.9M & - \\
	 \midrule
	 \multirow{4}{*}{16M} & 4M & 4.0M & 5.1M & 6.2M \\
	 & Objects365 & 0.58M  & - & 2.0M \\ 
	 & OpenImages & 1.7M  & - & 4.2M \\
	 & CC-12M & 11.1M & 11.1M & - \\
	 \bottomrule
	\end{tabular}
\end{table}

We compare X-VLM with existing approaches at two settings, as listed in Table~\ref{tbl:data}. We refer to them as the 4M setting and 16M setting respectively. Following UNITER~\cite{chen2020uniter} and other existing work, we prepare our pre-training data using two in-domain datasets, COCO~\cite{lin2014microsoft} and Visual Genome (VG)~\cite{krishna2016visual}, and two out-of-domain datasets, SBU Captions~\cite{ordonez2011im2text} and Conceptual Captions (CC)~\cite{sharma2018conceptual}.

In the 4M setting, we utilize image annotations only from COCO and VG, which contain 2.5M object annotations and 3.7M region annotations. Note that BUTD, the most widely used object detector, is trained on the same set of object annotations. The existing methods of only learning image-text alignments also utilize the region annotations of VG under the assumption that region descriptions can describe the whole images. In contrast, we take the object labels as text descriptions of objects, and re-formulate the image annotations so that an image has multiple boxes and each box is associated with a text. The text describes the visual concept in the box, which can be an object, a region, or the image itself.

In the 16M setting, we exploit a much noisier Conceptual 12M dataset (CC-12M)~\cite{changpinyo2021conceptual} following ALBEF~\cite{li2021align}. We additionally exploit Objects365~\cite{shao2019objects365} and OpenImages~\cite{kuznetsova2018open} following VinVL~\cite{zhang2021vinvl}.

Since most downstream V+L tasks are built on top of COCO and VG,  we exclude all images that also appear in the validation and test sets of downstream tasks to avoid information leak. We also exclude all co-occurring Flickr30K~\cite{plummer2015flickr30k} images via URL matching, because COCO and VG are from Flickr, and there are some overlaps.

\subsection{Implementation Details}
\label{sec:details}

The image encoder of X-VLM is vision transformer~\cite{dosovitskiy2020image}, which is initialized with Swin Transformer$_\mathrm{base}$~\cite{liu2021swin}. The text encoder and the cross-modal encoder consist of six transformer layers respectively. The text encoder is initialized using the first six layers of BERT$_\mathrm{base}$~\cite{devlin2019bert}, and the cross-modal encoder is initialized using the last six layers. In total, X-VLM has 215.6M parameters for pre-training.

X-VLM takes images of resolution of $224\times224$ as input. For text input, we set the maximum number of tokens to 30. During fine-tuning, we increase the image resolution to $384\times384$ and interpolate the positional embeddings of image patches following \citet{dosovitskiy2020image}.

We apply mixed precision for pre-training. In the 4M setting, we train the model for 200K steps on 8 NVIDIA A100 GPUs and the batch size is set to 1024, which tasks $\sim3.5$ days. In the 16M setting, we train the model on 24 GPUs with a batch size of 3072. We sample the data by making half of the images in a batch containing bounding box annotations. We use the AdamW~\cite{loshchilov2018decoupled} optimizer with a weight decay of 0.02. The learning rate is warmed-up to $1e^{-4}$ from $1e^{-5}$ in the first 2500 steps and decayed to $1e^{-5}$ following a linear schedule.

\begin{table*}[!t]
	\caption
	{
		Image-text retrieval results on MSCOCO and Flickr30K datasets. IR: Image Retrieval and TR: Text Retrieval. We compute Recall@K with K = 1, 5, 10, as the evaluation metric. Zero-shot retrieval results are given in Appendix~\ref{app:zeroshot}. 
	}
	\label{tbl:retrieval}
    \small
	\centering	
	\begin{tabular}	{l  c c |  c  c | c  c  c  c  c  c }
		\toprule	 	
	 \multirow{2}{*}{Method} & \multirow{2}{*}{\# Params} & \# Pre-train & \multicolumn{2}{c|}{MSCOCO (5K test set)} & \multicolumn{2}{c}{Flickr30K (1K test set)} \\
	 & & ~~Images &  TR & IR & TR & IR\\
	 \midrule
	& & & R@1/R@5/R@10 & R@1/R@5/R@10 & R@1/R@5/R@10 & R@1/R@5/R@10\\
	UNITER$_\mathrm{large}$ & 300M & 4M & 65.7 / 88.6 / 93.8 & 52.9 / 79.9 / 88.0 & 87.3 / 98.0 / 99.2 & 75.6 / 94.1 / 96.8 \\
    METER-Swin & 380M & 4M & 73.0 / 92.0 / 96.3 & 54.9 / 81.4 / 89.3 & 92.4 / 99.0 / 99.5 & 79.0 / 95.6 / 98.0 \\
	ALBEF & 210M & 4M & 73.1 / 91.4 / 96.0 & 56.8 / 81.5 / 89.2 & 94.3 / 99.4 / 99.8  & 82.8 / 96.7 / 98.4 \\
    METER-CLIP & 380M & 4M & 76.2 / 93.2 / 96.8 & 57.1 / 82.7 / 90.1 & 94.3 / 99.6 / 99.9 & 82.2 / 96.3 / 98.4 \\ 
	VinVL$_\mathrm{large}$ & 550M & 5.6M & 75.4 / 92.9 / 96.2 & 58.8 / 83.5 / 90.3 & -  & - \\
	ALIGN & 490M & 1.8B & 77.0 / 93.5 / 96.9 & 59.9 / 83.3 / 89.8 & 95.3 / 99.8 / 100.0 & 84.9 / 97.4 / 98.6 \\
	ALBEF & 210M & 14M & 77.6 / 94.3 / 97.2 & 60.7 / 84.3 / 90.5 & 95.9 / 99.8 / 100.0  & 85.6 / \textbf{97.5} / \textbf{98.9} \\
	\midrule
	X-VLM & 216M & 4M & 80.4 / 95.5 / \textbf{98.2} & 63.1 / 85.7 / \textbf{91.6} & 96.8 / 99.8 / 100.0 & 86.1 / 97.4 / 98.7 \\
	X-VLM & 216M & 16M & \textbf{81.2} / \textbf{95.6} / \textbf{98.2} & \textbf{63.4} / \textbf{85.8} / 91.5 & \textbf{97.1} / \textbf{100.0} / 100.0 & \textbf{86.9} / 97.3 / 98.7 \\
		\bottomrule
	\end{tabular}
\end{table*}

\begin{table*}[!t]
	\caption
	{
		Results on downstream V+L tasks, including visual reasoning (VQA and NLVR2), visual grounding (RefCOCO+), and image caption generation (COCO Caption). RefCOCO+ scores with $^*$ are evaluated in the weakly-supervised setting. COCO Captioning scores with $^+$ are models optimized with CIDEr for the second stage of fine-tuning. 
 	}
	\label{tbl:results}
    \small
	\centering	
	\begin{tabular}	{l   |  c  c  c  c  c c c c c}
	\toprule
	 \multirow{2}{*}{Method} & \multicolumn{2}{c}{VQA} & \multicolumn{2}{c}{NLVR2} & \multicolumn{3}{c}{RefCOCO+} & \multicolumn{2}{c}{COCO Caption} \\
	  & test-dev & test-std & dev & test-P & val$^d$ & testA$^d$ & testB$^d$ & BLEU@4 & CIDEr \\
	  \midrule
	  ViLBERT & 70.55 & 70.92 & - & - & 72.34 & 78.52 & 62.61 & - & - \\
	  VL-BERT & 71.16 & - &  - & - & 72.59 & 78.57 & 62.30 & - & - \\
	   VILLA & 73.59 & 73.67 & 78.39 & 79.30 & 76.05 & 81.65 & 65.70 & - & - \\
	  SOHO & 73.25 & 73.47 & 76.37 & 77.32 & - & - & - & - & - \\
	   E2E-VLP & 73.25 & 73.67 & 77.25 & 77.96 & - & - & - & 36.2 & 117.3 \\
	   	KD-VLP & 74.20 & 74.31 & 77.36 & 77.78 & - & - & - & - & - \\
  	  UNITER$_\mathrm{large}$ & 73.82 & 74.02 & 79.12 & 79.98 & 75.90 & 81.45 & 66.70 & - & - \\
     ALBEF(4M) & 74.54 & 74.70 & 80.24 & 80.50 & - & - & - & - & - \\
   	  ALBEF(14M) & 75.84 & 76.04 & 82.55 & 83.14 & 58.46$^*$ & 65.89$^*$ & 46.25$^*$ & - & -\\
   	       METER-Swin & 76.43 & 76.42 & 82.23 & 82.47 & - & - & - & - & - \\
      	   VinVL$_\mathrm{large}$(5.6M) & 76.52 & 76.60 & 82.67 & 83.98 & - & - & - & 41.0$^+$ & \textbf{140.9}$^+$ \\
   	      METER-CLIP & 77.68 & 77.64 & 82.33 & 83.05 & - & - & - & - & - \\
   	     SimVLM$_\mathrm{base}$(1.8B) &  77.87 & 78.14 & 81.72 & 81.77 & - & - & - & 39.0 & \textbf{134.8} \\
	  \midrule
	  X-VLM(4M) & 78.07 & 78.09 & 84.16 & 84.21 & 80.17 & 86.36 & 71.00 & 39.8 / \textbf{41.3}$^+$ & 133.1 / 140.8$^+$ \\
	  X-VLM(16M) & \textbf{78.22} & \textbf{78.37} & \textbf{84.41} & \textbf{84.76} & \textbf{84.51} & \textbf{89.00} & \textbf{76.91} & \textbf{39.9} / 41.0$^+$ & 134.0 / 140.3$^+$ \\
  	  	\bottomrule
  	  
	\end{tabular}
\end{table*}

\subsection{Downstream Tasks}

We adapt X-VLM to five downstream V+L tasks. We follow the settings in the previous work on fine-tuning (see Appendix~\ref{app:details}). Note that we have cleaned the pre-training datasets to avoid data leaks since downstream V+L tasks have overlaps in images with COCO and Visual Genome.

\noindent \textbf{Image-Text Retrieval} There are two subtasks: text retrieval (TR) and image retrieval (IR). We evaluate X-VLM on MSCOCO and Flickr30K~\cite{plummer2015flickr30k} datasets. We adopt the widely used Karpathy split~\cite{karpathy2015deep} for both datasets. We optimize $\mathcal{L}_\mathrm{cl}$ and $\mathcal{L}_\mathrm{match}$ and fine-tune the model for 10 epochs. In inference, we first compute $s(I,T)$ for all images and texts, and then take the top-$k$ candidates and calculate $\Vec{p}^\textrm{match}(I,T)$ for ranking. Following ALBEF, $k$ is set to 256 for MSCOCO and 128 for Flickr30K.

\noindent \textbf{Visual Question Answering}~\cite{goyal2017making} It requires the model to predict an answer given an image and a question. Following the previous work~\cite{cho2021unifying, li2021align}, we use a six-layer Transformer decoder to generate answers based on the outputs of the cross-modal encoder. We fine-tune the model for 10 epochs. During inference, we constrain the decoder to only generate from the 3,129 candidate answers to make a fair comparison with existing methods.

\noindent \textbf{Natural Language for Visual Reasoning} (NLVR2~\cite{suhr2018corpus}) The task lets the model determine whether a text describes the relations between two images. Following ALBEF, we extend the cross-modal encoder to enable reasoning over two images, and perform an additional pre-training step for one epoch using the 4M images: given two images and a text, the model assigns the text to either the first image, the second image, or none of them. Then, we fine-tune the model for 10 epochs.

\noindent \textbf{Visual Grounding} The task (RefCOCO+~\cite{yu2016modeling}) aims to locate the region in an image that corresponds to a specific text description. Previous approaches formulate grounding as a ranking task by relying on the region proposals provided by pre-trained object detectors~\cite{lu2019vilbert, su2019vl, chen2020uniter, gan2020large}. In contrast, X-VLM is able to directly predict the bounding boxes of the target regions given images and text descriptions. We also evaluate X-VLM on a weakly-supervised setting, proposed by ALBEF, in which case only image-text pairs are available, and thus we fine-tune X-VLM using $\mathcal{L}_\mathrm{cl}$ and $\mathcal{L}_\mathrm{match}$;

\noindent \textbf{Image Captioning} The task requires a model to generate textual descriptions of input images. We evaluate X-VLM on the COCO Captioning dataset~\cite{chen2015microsoft}. We report BLEU-4 and CIDEr scores on the Karparthy test split. To apply X-VLM for captioning, we do not need to add a decoder. Instead, we simply adapt X-VLM to a multi-modal decoder. Specifically, we train X-VLM with language modeling loss for one epoch on 4M data. Then, we fine-tune it on the COCO Captioning dataset. Additionally, following VinVL, we also report the results after applying CIDEr optimization~\cite{rennie2017self} for the second stage of fine-tuning, which are denoted with $^+$.

\begin{figure*}[ht]
\centering
\centerline{\includegraphics[width=2\columnwidth]{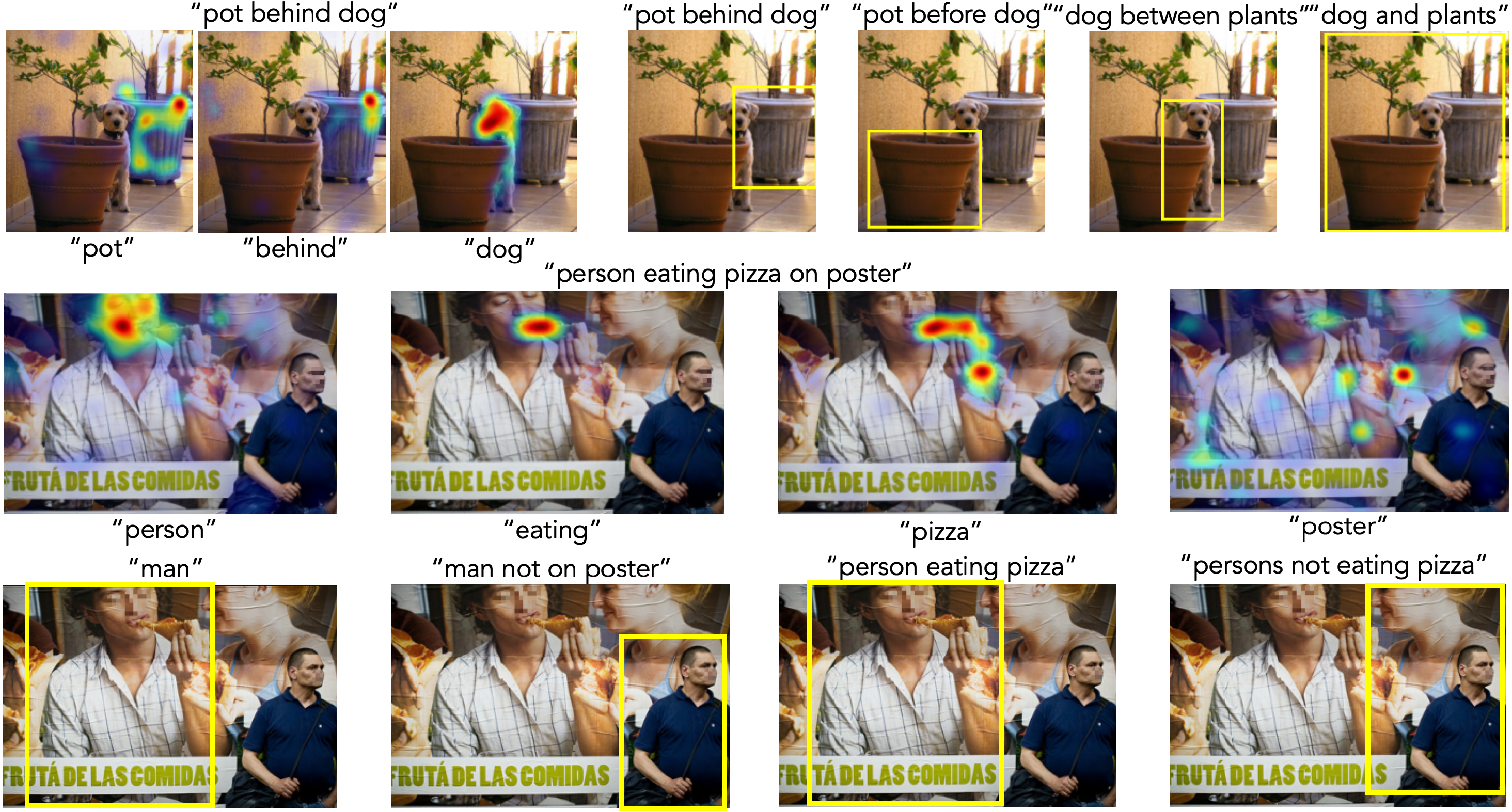}}
\caption{Grad-CAM visualization and bounding box prediction on unseen images. X-VLM predicts correct regions even though the textual descriptions only differ in a single word. X-VLM can also align each word in the text to the corresponding image region. Appendix \ref{app:pic} gives more examples, showing X-VLM's superior ability of multi-grained vision language alignments. }
\label{Fig:refcoco_all}
\end{figure*}

\subsection{Results on Image-Text Retrieval}
Table~\ref{tbl:retrieval} compares X-VLM with SoTA approaches on MSCOCO and Flickr30K, which are based on either object-centric features (i.e. UNITER and VinVL) or overall image features (i.e. ALIGN, METER, and ALBEF). ALIGN~\cite{jia2021scaling} is a dual-encoder model similar to CLIP~\cite{radford2021learning} specially for image-text retrieval tasks, which is trained on in-house 1.8B image-text pairs. Other VLMs, including our approach, for more general purposes, have a cross-modal encoder and thus use the output of the cross-modal encoder for ranking.

Even though existing approaches either have more parameters or more training data, X-VLM under the 4M setting outperforms all the previous methods by a large margin, achieving new SoTA results. Specifically, X-VLM(4M) which learns multi-grained vision language alignments outperforms VinVL which is based on object-centric features. In contrast, ALBEF which learns only image-text alignments outperforms VinVL only when increasing the training data to 14M. Compared to METER-Swin~\cite{dou2021empirical} which also uses Swin Transformer as the image encoder, X-VLM has better performance. Furthermore, even though X-VLM(4M) has already achieved very high performance on the image-text retrieval tasks, we still obtain improvements on R@1 when increasing the training instances to 16M. Additionally, Appendix \ref{app:zeroshot} shows that when increasing the training data to 16M, X-VLM obtains substantial improvements on zero-shot image-text retrieval. Moreover, X-VLM also outperforms ALIGN on zero-shot MSCOCO by a large margin.

Additionally, METER provides an empirical study of VLMs and shows that the vision backbone (or parameter initialization) is important for the model performance. From Swin Transformer to CLIP-ViT, METER improves significantly on both retrieval and VQA (Table~\ref{tbl:retrieval} and~\ref{tbl:results}). We also have some preliminary observations and leave detailed studies of different backbones of X-VLM for future work.

\begin{table*}[t]
	\caption
	{
		Ablation study results. Models w/o object and w/o region are ablated variants where the model is training without concepts of object and region respectively. Model w/o bbox loss is the variant where bounding box prediction is ablated. Model w/o all represents that all the above components are ablated. 
	}
	\label{tbl:ablation}
    \small
	\centering	
	\begin{tabular}	{l | r | c  c  c  c  c c r r }
		\toprule	 	
	 & \multicolumn{1}{c|}{Meta-Sum} & \multicolumn{2}{c}{MSCOCO} & \multicolumn{2}{c}{Flickr30K} & VQA & NLVR$^2$ & \multicolumn{2}{c}{RefCOCO+} \\
	 & & TR & IR & TR & IR & test-dev & test-P & \multicolumn{1}{c}{testA$^d$} & \multicolumn{1}{c}{testB$^d$} \\
	\midrule
    X-VLM & \textbf{605.0} &  \textbf{78.8} & \textbf{60.6} & \textbf{96.0} & \textbf{84.1} & 76.20 & 82.42 & 72.07 & 54.84 \\  
	\midrule
    w/o object & 603.5 & 77.4  & 60.4   & 95.0  & 83.7 & 75.87  & 82.10 & \textbf{73.37} & \textbf{55.69} \\  
    w/o region & 596.0 & 76.8 & 60.2  & \textbf{96.0} & 83.6 & 75.84  & 82.20 & 70.73 & 50.60 \\ 
    w/o bbox loss & 594.9 & 77.5 & 60.2 & 95.7 & 83.5 & \textbf{76.77} & 81.49 & 69.32 & 50.38 \\
    w/o all & 580.6 & 74.5 & 57.9 & 95.6 & 82.8 & 74.90  & 80.70 & 67.79 & 46.43 \\
		\bottomrule
	\end{tabular}
\end{table*}

\subsection{Results on Visual Reasoning}

Table~\ref{tbl:results} shows experimental results on visual reasoning (VQA and NLVR$^2$). First, though ALBEF(14M) outperforms VinVL on image-text retrieval, the coarse-grained approaches such as SOHO, METER-Swin, and ALBEF, all have worse performances than VinVL in visual reasoning tasks, except that METER-CLIP and SimVLM outperform VinVL on VQA. Besides, VinVL also substantially outperforms previous methods that rely on object detectors to learn both object-level and image-level alignments, such as E2E-VLP and KD-VLP.

Nevertheless, X-VLM(4M) with moderate model size and pre-trained on fewer instances outperforms VinVL. Specifically, X-VLM(4M) achieves absolute improvements of $1.52\%$ on VQA and $0.86\%$ on NLVR2 (average on metrics) over VinVL. Meanwhile, as reported in \citet{li2021align}, X-VLM, which encodes images without an object detection process, enjoys $\sim10$ times faster inference speed than VinVL. The results indicate that our approach of X-VLM is both effective and efficient. X-VLM also outperforms SimVLM$_\mathrm{base}$ which is pre-trained on in-house 1.8B data, especially on NLVR2.

\subsection{Results on Visual Grounding}

Table~\ref{tbl:results} reports the performance of X-VLM on RefCOCO+. X-VLM(4M) achieves absolute improvements of $4.5\%$ compared to UNITER. As aforementioned, previous approaches formulate grounding as a ranking task by relying on the region proposals provided by object detectors. In contrast, X-VLM is able to directly predict the target boxes, which is much simpler and more efficient. Furthermore, X-VLM for general V+L purposes outperforms MDETR~\cite{kamath2021mdetr} specialized for visual grounding tasks. X-VLM(4M) using the same set of image annotations achieves absolute improvements of $1.1\%$ (average on metrics), compared to MDETR.

We also evaluate X-VLM in the weakly-supervised setting, proposed by ALBEF. X-VLM(4M) obtains 68.46/76.53/57.09 for val$^d$/testA$^d$/testB$^d$ respectively, achieving an absolute improvement of $10.5\%$ (average on metrics) compared to ALBEF(14M). When increasing pre-training images to 16M, X-VLM obtains 77.26/84.11/67.13.

Figure~\ref{Fig:refcoco_all} provides a few examples of images from the test set of RefCOCO+. For the supervised setting, we show the bounding boxes predicted by X-VLM given the text descriptions. For the weakly-supervised setting, following ALBEF, we provide the Grad-CAM visualization, which uses the cross-attention maps in the fourth layer of the cross-modal encoder. The visualization examples show that X-VLM has a strong ability of cross-modal understanding. It successfully predicts the correct regions in images, even though the text descriptions only differ in a single word. Furthermore, X-VLM can align each word in the text to the corresponding image region. We provide more examples in Appendix~\ref{app:pic}, showing X-VLM's superior performance in vision language alignment.

\subsection{Results on Image Captioning}

We show that X-VLM, usually considered as an ``encoder-only'' model, has comparable performance with SoTA generative methods on image caption generation, as indicated in Table~\ref{tbl:results}. Specifically, X-VLM pre-trained on 16M instances performs similarly to SimVLM which uses not only 1.8B in-house image-text pairs but also a large-scale text corpus. Besides, we observe that CIDEr optimization largely boosts the CIDEr scores. X-VLM in moderate model size also has comparable performance to VinVL$_\mathrm{large}$.

\subsection{Ablation Study}

We also conduct an in-depth ablation study to investigate the role of different components in the X-VLM, as shown in Table~\ref{tbl:ablation}. All compared model variants are trained on 4M images for 80K steps with a batch size of 3072 to ensure a fair comparison. We use Recall@1 as an evaluation measure in the retrieval tasks and Meta-Sum as a general measure. We report RefCOCO+ evaluation results in the weakly-supervised setting.

First, we evaluate the effectiveness of visual concepts in different granularities, i.e. w/o object and w/o region. The results show that training without either of them hurts the performance, demonstrating the necessity of learning multi-grained alignments. Besides, we can observe that w/o region makes the performance drop more drastically than w/o object. Furthermore, the ablation study shows that bounding box prediction is a critical component of X-VLM, as w/o bbox loss leads to the lowest Meta-Sum. We also report the results of `w/o all' where all the above components are ablated. Though in the 4M setting, only 210K images have dense annotations, X-VLM can leverage the data to learn multi-grained vision language alignment and substantially improve the performances in the downstream V+L tasks (Meta-Sum from 580.6 to 605.2).

\section{Conclusion and Discussion}

In this paper, we have proposed X-VLM, a strong and efficient approach to perform multi-grained vision language pre-training. Training of the model is driven by locating visual concepts in the image given the associated texts and aligning texts with relevant visual concepts, where the alignments are in multi-granularity. We have pre-trained X-VLM with 4M and 16M images, which are of moderate size. Also, X-VLM only consists of 216M parameters. These choices are made because we want to make our experiments as ``green''~\cite{schwartz2020green,xu2021survey} as possible and be accessible to a larger group of people. Experiments on downstream V+L tasks, including image-text retrieval, visual reasoning, visual grounding, and image caption generation have shown that X-VLM outperforms the existing methods which could be larger and/or pre-trained on more data. As suggested by the comparison between X-VLM(4M) and X-VLM(16M), adding more pre-training datasets will probably lead to further performance improvements. As for applications, X-VLM has shown better performance in understanding fine-grained vision language alignments. For example, it can generate image captions probably having more object details, which makes it a better choice to help people with disability in vision to understand images. On the other hand, X-VLM in moderate model size is also easier to deploy.

\section*{Acknowledgements}
We would like to thank Wangchunshu Zhou, Wenguan Huang, and Xiu-jun Li at ByteDance for their generous assistance in data collection and insightful comments in technical discussions.


\bibliography{main}
\bibliographystyle{icml2022}

\newpage
\appendix
\onecolumn

\section{Appendix}
\subsection{Statistics of Object and Region Annotations}
\label{app:ann}

\begin{table}[ht]
	\caption
	{
	Statistics of annotations used in the pre-training.
	}
	\label{tbl:ann}
    \small
	\centering	
	\begin{tabular}	{l | l | l |  l | l }
	\toprule
	 Dataset & \# Images  & \# Captions & \# Objects & \# Regions \\
\midrule
	 COCO & 0.11M & 0.55M & 0.45M & - \\
	 VG & 0.10M & - & 2.0M & 3.7M \\
	 Objects365 & 0.58M  & - & 2.0M & - \\ 
	 OpenImages & 1.7M  & - & 4.2M & - \\
	 \bottomrule
	\end{tabular}
\end{table}

Table \ref{tbl:ann} gives statistics of object and region annotations of each dataset. Only the Visual Genome dataset contains region annotations. Besides, the OpenImages dataset offers some relationship annotations, indicating pairs of objects in particular relations (e.g. "woman playing guitar", "beer on table"), object properties (e.g. "table is wooden"), and human actions (e.g. "woman is jumping"), which can also be viewed as region annotations.

Note that we filtered out some samples because of: 1) invalid annotations (e.g. negative values for bounding boxes or boxes being outside of the images); 2) boxes being too small ($<$ 1\%); 3) highly overlapped textual descriptions of regions ($>$ 75\%), etc. After pre-processing, we keep: for example, COCO objects 446,873 (from 859,999), VG objects 2,043,927 (from 3,802,349), VG regions 3,699,598 (from 5,402,953).

\subsection{Implementation Details of Downstream Tasks}
\label{app:details}

We follow the settings in existing methods for fine-tuning. 
We describe how we implement fine-tuning on the downstream V+L tasks, and we also provide our fine-tuning scripts for more details. Note that we have cleaned our pre-training datasets to avoid data leaks since downstream V+L tasks have overlaps in images with COCO and Visual Genome.

\noindent \textbf{Image-Text Retrieval} We evaluate X-VLM on MSCOCO and Flickr30K~\cite{plummer2015flickr30k} benchmarks. We adopt the widely used Karpathy split~\cite{karpathy2015deep} for both datasets. We optimize $\mathcal{L}_\mathrm{cl}$ and $\mathcal{L}_\mathrm{match}$ for fine-tuning. Since there are multiple ground-truth texts associated with each image in the datasets, we change the ground-truth similarity of contrastive learning, $\Vec{y}^\mathrm{v2t}(I)$ and $\Vec{y}^\mathrm{t2v}(T)$, to consider multiple positives, where each positive example has a probability of $\frac{1}{\#\mathrm{positives}}$. We fine-tune the model for 10 epochs. During inference, we first compute $s(I,T)$ for all images and texts. Then we take the top-$k$ candidates and pass them into the cross-modal encoder to calculate $\Vec{p}^\textrm{match}(I,T)$ for ranking. Following ALBEF, $k$ is set to 256 for MSCOCO and 128 for Flickr30K.

\noindent \textbf{Visual Question Answering} (VQA~\cite{goyal2017making}) Following existing methods~\cite{tan2019lxmert, chen2020uniter, li2021align}, we use both train and validation sets for training, and include additional question-answer pairs from Visual Genome. The VQA model contains a 6-layer transformer-based decoder to generate answers based on the outputs of the cross-modal encoder following previous work~\cite{cho2021unifying, li2021align}. The decoder is initialized using the pre-trained weights from the cross-modal encoder. Then, the model is fine-tuned by optimizing the auto-regressive loss for 10 epochs. During inference, we constrain the decoder to only generate from the 3,129 candidate answers \footnote{There is a NULL answer. Thus, the actual number of candidate answers is 3,128.} to make a fair comparison with existing methods.

\noindent \textbf{Natural Language for Visual Reasoning}  (NLVR2~\cite{suhr2018corpus}) Since the task asks the model to distinguish whether a text describes a pair of images, we follow ALBEF to extend the cross-modal encoder to enable reasoning over two images. We also perform an additional pre-training step for 1 epoch using the 4M images: given a pair of images and a text, the model needs to assign the text to either the first image, the second image, or none of them. Then, we fine-tune the model for 10 epochs.

\noindent \textbf{Visual Grounding} The task aims to locate the region in an image that corresponds to a specific text description (RefCOCO+~\cite{yu2016modeling}). We evaluate our approach in both supervised and weakly-supervised settings. The latter is proposed by ALBEF. In the supervised setting with bounding box annotations, we perform an additional pre-training step for one epoch using $\mathcal{L}_\mathrm{bbox}$ only. Then, we fine-tune the model for 10 epochs. In the weakly-supervised setting where only image-text pairs are available, we fine-tune the model using $\mathcal{L}_\mathrm{cl}$ and $\mathcal{L}_\mathrm{match}$ for 5 epochs. During inference, following ALBEF, we apply Grad-CAM~\cite{selvaraju2017grad} to acquire heatmaps and use them to rank the detected proposals provided by~\cite{yu2018mattnet}.

\noindent \textbf{Image Captioning} The task requires a model to generate textual descriptions of input images. We evaluate X-VLM on the COCO Captioning dataset~\cite{chen2015microsoft}. We report BLEU-4 and CIDEr scores on the Karparthy test split. To apply X-VLM for captioning, we do not need to add a decoder. Instead, we simply adapt X-VLM to a multi-modal decoder. Specifically, we train X-VLM with language modeling loss for one epoch on 4M data. Then, we fine-tune it on the COCO Captioning dataset with naive cross-entropy loss for five epochs. Additionally, following VinVL, we also report the results after applying CIDEr optimization~\cite{rennie2017self} for the second stage of fine-tuning which takes another five epochs.

\subsection{Zero-Shot Image-Text Retrieval Results}
\label{app:zeroshot}

\begin{table*}[h]
	\caption
	{
		Zero-shot results on MSCOCO and Flickr30K datasets. IR: Image Retrieval and TR: Text Retrieval. 
	}
	\label{tbl:zeroretrieval}
    \small
	\centering	
	\begin{tabular}	{l  c c |  c  c | c  c  c  c  c  c }
		\toprule	 	
	 \multirow{2}{*}{Method} & \multirow{2}{*}{\# Params} & \# Pre-train & \multicolumn{2}{c|}{MSCOCO (5K test set)} & \multicolumn{2}{c}{Flickr30K (1K test set)} \\
	 & & ~~Images &  TR & IR & TR & IR\\
	 \midrule
	& & & R@1/R@5/R@10 & R@1/R@5/R@10 & R@1/R@5/R@10 & R@1/R@5/R@10\\
	CLIP & $\sim$100M & 400M & 58.4 / 81.5 / 88.1 & 37.8 / 62.4 / 72.2 & 88.0 / \textbf{98.7} / 99.4 & 68.7 / 90.6 / 95.2 \\ 
	ALIGN & 490M & 1.8B & 58.6 / 83.0 / 89.7 & 45.6 / 69.8 / 78.6 & \textbf{88.6} / \textbf{98.7} / \textbf{99.7} & \textbf{75.7} / 93.8 / 96.8 \\
	\midrule
	X-VLM & 216M & 4M &  70.8 / 92.1 / 96.5 & 55.6 / 82.7 / 90.0 & 85.3 / 97.8 / 99.6 & 71.9 / 93.3 / 96.4 \\
	X-VLM & 216M & 16M & \textbf{71.6} / \textbf{93.1} / \textbf{97.0} & \textbf{56.1} / \textbf{83.0} / \textbf{89.8} & 87.7 / 98.6 / 99.6 & 74.9 / \textbf{94.4} / \textbf{97.1} \\
		\bottomrule
	\end{tabular}
\end{table*}

Table \ref{tbl:zeroretrieval} shows zero-shot image-text retrieval results and compares X-VLM with the dual encoder SoTAs (CLIP and ALIGN) which are pre-trained using only the retrieval objective. We can observe that though X-VLM is pre-trained using the combination of different objectives, it still has very competitive results on zero-shot retrieval tasks.

\subsection{Case Study}
\label{app:pic}

\begin{figure*}[t]
\centering
\centerline{\includegraphics[width=1.0\columnwidth]{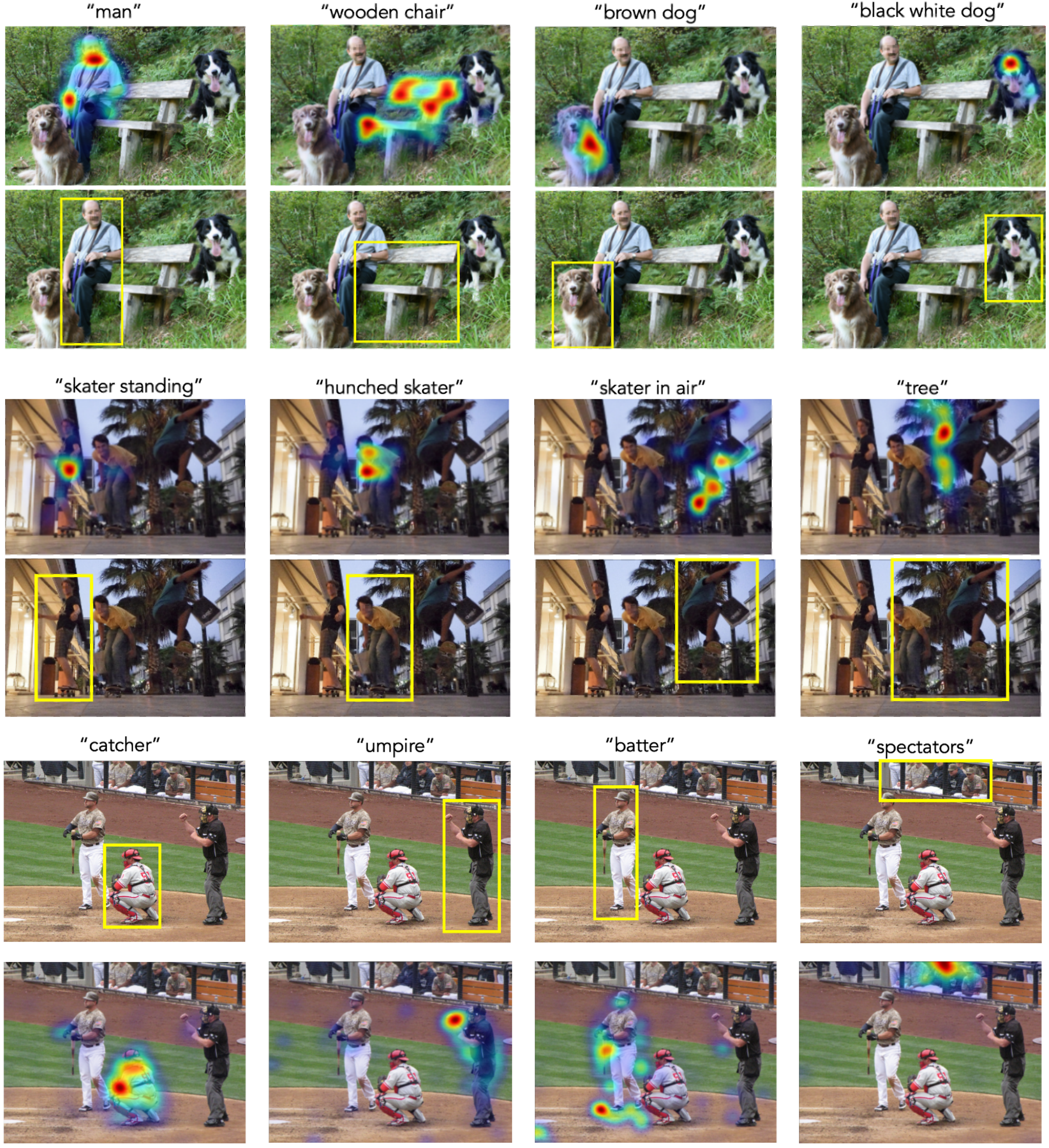}}
\caption{Locating visual concepts in unseen images given text descriptions. Since Grad-CAM gives visualizations each corresponds to an individual word, we only show the visualization of the subject word, e.g. "dog" for "brown dog". }
\label{Fig:app_visualization_for_detection}
\end{figure*}

\begin{figure*}[t]
\centering
\centerline{\includegraphics[width=1.0\columnwidth]{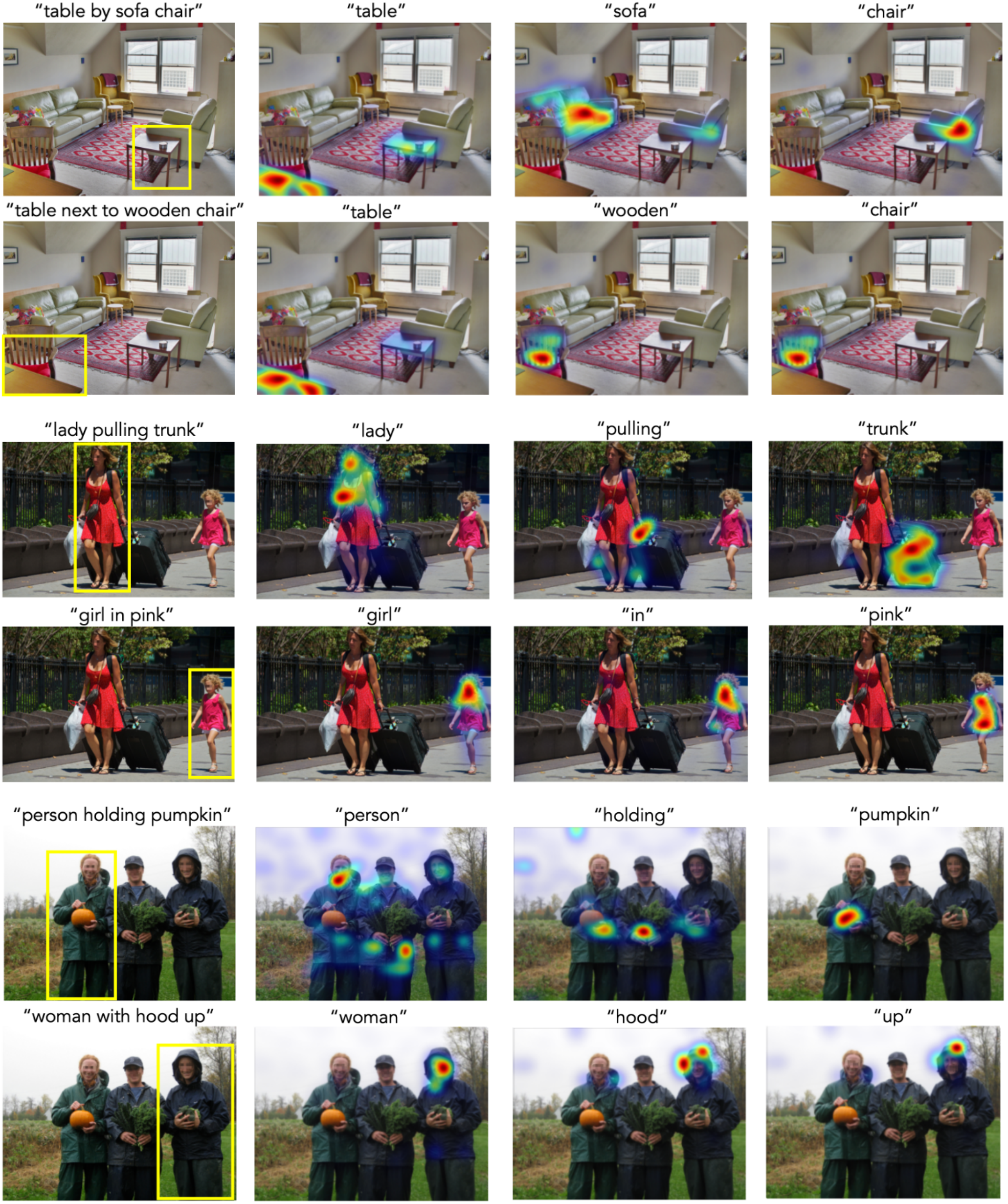}}
\caption{Bounding box prediction and per-word visualization on unseen images. It shows that X-VLM can also align concepts like ``pulling'' and ``holding'' to the corresponding regions in the images. }
\label{Fig:app_per_word_visualization}
\end{figure*}

Figure \ref{Fig:app_visualization_for_detection} and \ref{Fig:app_per_word_visualization} provide visualizations of some images from the test set of RefCOCO+. We show the bounding boxes predicted by X-VLM given the text descriptions. For the weakly-supervised setting, we provide the Grad-CAM visualization which uses the cross-attention maps in the fourth layer of the cross-modal encoder. We can observe that in both settings X-VLM can predict correct regions even though the textual descriptions only differ in a single word. X-VLM can also align each word in the text to the corresponding image region, showing X-VLM's superior ability of multi-grained vision language alignments.

\end{document}